\documentclass{article}
\usepackage{ijcai19}

\usepackage{times}
\usepackage{soul}
\usepackage{url}
\usepackage[hidelinks]{hyperref}
\usepackage[utf8]{inputenc}
\usepackage[small]{caption}
\usepackage{graphicx}
\usepackage{amsmath}
\usepackage{booktabs}
\usepackage{algorithm}
\usepackage{algorithmic}
\usepackage{subcaption}
\usepackage{complexity}
\usepackage{multicol}
\usepackage{booktabs}
\usepackage{scalefnt}
\usepackage{multirow}
\usepackage{comment}
\usepackage{amsmath}
\usepackage{amsthm}
\usepackage{array}
\usepackage[framemethod=tikz]{mdframed}
\usepackage{tikz}
\usepackage{wrapfig}

\newtheorem{definition}{Definition}
\newtheorem{example}{Example}

\newcommand{\cor}[1]{
  \cellcolor{orange!40}
}

\usepackage{todonotes}

\providecommand{\tuple}[1]{\ensuremath{\langle #1\rangle}}

\usepackage{stmaryrd}
\providecommand{\apply}[1]{\ensuremath{\llbracket #1\rrbracket}}

\DeclareMathOperator{\successor}{succ}
\DeclareMathOperator{\predecessor}{pred}

\title{Procedural Generation of Initial States of Sokoban}

\author{
Dâmaris S. Bento,\textsuperscript{\rm 1} 
André G. Pereira\textsuperscript{\rm 2} \And 
Levi H. S. Lelis\textsuperscript{\rm 1} 
\affiliations
$^1$Universidade Federal de Viçosa, Brazil\\
$^2$Universidade Federal do Rio Grande do Sul, Brazil\\
\emails
\{damaris.bento, levi.lelis\}@ufv.br,
agpereira@inf.ufrgs.br
}

\begin{document}

\maketitle

\begin{abstract}
Procedural generation of initial states of state-space search problems have applications in human and machine learning as well as in the evaluation of planning systems. In this paper we deal with the task of generating hard and solvable initial states of Sokoban puzzles. We propose hardness metrics based on pattern database heuristics and the use of novelty to improve the exploration of search methods in the task of generating initial states. 
We then present a system called $\beta$ that uses our hardness metrics and novelty to generate initial states. Experiments show that $\beta$ is able to generate initial states that are harder to solve by a specialized solver than those designed by human experts. %$\beta$ outperforms current methods by generating initial states with an order of magnitude more boxes than the states generated by existing approaches.
\end{abstract}

\section{Introduction}

The development of search solvers for state-space search problems is a traditional area of research in artificial intelligence (AI), where notable progress has been made (e.g., \cite{CulbersonSchaeffer1996,Junghanns2001,Hoffmann2001,Holte2016}). By contrast, systems for generating problems have achieved limited success. Generation systems are often able to generate only small and easy problems, thus limiting their applicability.  

Systems able to automatically generate solvable state-space search problems have applications in human and machine learning. These systems can be used to generate problems from which humans~\cite{Anderson85} and systems~\cite{Arfaee2011,Lelis2016,Racaniere2017,Orseau2018} 
can learn. Automatically generated problems are also used to evaluate planning systems. Problems used in the International Planning Competition are usually generated using domain knowledge and their solvability verified with specialized solvers. 
The generation of solvable and hard problems can be challenging, since it could be \PSPACE-complete to decide if a problem is solvable. Moreover, it is unclear what makes a problem hard in practice.

In this paper we focus on the task of generating hard and solvable initial states of Sokoban. %, a map-based puzzle. 
We introduce hardness metrics based on pattern databases (PDBs)~\cite{CulbersonSchaeffer1996} motivated by hardness metrics known to correlate with the time required by humans to solve problems~\cite{Jaruvsek2010}. We also propose the use of 
%the state-space structural information through the concept of 
novelty~\cite{Lipovetzky2012} to improve the exploration in the task of generating initial states. Finally, we introduce a system called $\beta$ that performs a search guided by our metrics of hardness and novelty, and that also uses our metrics to select hard and solvable initial states.

%We use the Sokoban to evaluate $\beta$ for several reasons. 
Although $\beta$ is in principle general and applicable to other problem domains, we focus on Sokoban for several reasons. 
First, 
Sokoban is a \PSPACE-complete problem~\cite{Culberson1999} and a traditional testbed for search methods. Second, in Sokoban most of the states in its space are unsolvable, thus making the problem of generating solvable initial states particularly challenging. Third, Sokoban has a community of expert designers who create hard problems and make them available online, thus allowing a comparison between human-designed initial states with initial states generated by $\beta$. 
%experts with those generated by $\beta$. 

Our empirical results show that $\beta$ is able to generate solvable initial states that are harder to solve by a specialized solver than those designed by human experts.
To the best of our knowledge, our work is 
%the first focusing on the generation of initial states for state-space search problems and 
the first to compare the hardness of initial states generated by a computer system with those designed by human experts. 
%xSokoban~\cite{xSokoban} is the standard benchmark used in research to evaluate specialized solvers and is one of the hardest sets of problems available. xSokoban allows us to compare compare the initial states designed by human 
%Our experiments also show that $\beta$ is able to substantially outperform existing methods for generating initial states of Sokoban in terms of the size of the problem instances generated.  
%$\beta$  generates 
%hard and solvable initial states with much larger state spaces, longer solutions, and 
%initial states with an order of magnitude more boxes than existing methods~\cite{Murase1996,Taylor2011,kartal2016a}. 

\paragraph{Contributions} We introduce computationally efficient hardness metrics and show empirically how these metrics can be used to guide a search algorithm and to select hard instances from a pool of options. Another contribution is to show  that novelty~\cite{Lipovetzky2012} is essential for generating a diverse pool of solvable instances from which hard ones can be selected. The combination of our metrics and novelty resulted in the first generative system for Sokoban with superhuman performance in terms of instance difficulty for a specialized solver. 

%Although we evaluate $\beta$ in Sokoban, in principle $\beta$ can be used to generate initial states of other domains. This is because our hardness metrics are general as they use domain-independent PDB heuristics~\cite{Edelkampecp01} and the exploration metric of novelty has been shown to be effective on a wide range of problems~\cite{Lipovetzky2017}. 
%can thus be used to evaluate the difficulty of other problem domains. 
%~\cite{Edelkampecp01}. 
%The exploration techniques can also be used for planning problems. In addition, Sokoban is one hardest planning problems and belongs to the class of transportation problems -- the most common class of planning problems. Thus our techniques are likely to present robust results for problems in this class . Our approach is general and addresses a significant problem with many applications.

\section{Related Work}

Related to our work are papers presenting methods for generating entire problems of Sokoban, and more generally, works dealing with automatic content generation. Also related to our work are learning systems that require the generation of a set of states from which one can learn a value or a heuristic function. Next, we briefly survey each of these related areas. 
 %some of these systems. 
%presenting methods for generating puzzles and games. 

%Previous work showed how to generate an entire Sokoban problem: initial state and maze~\cite{Murase1996,Taylor2011,kartal2016a}. 

\citeauthor{Murase1996}~\shortcite{Murase1996}, \citeauthor{Taylor2011}~\shortcite{Taylor2011}, and \citeauthor{kartal2016a}~\shortcite{kartal2016a} presented methods for generating complete Sokoban problems, where a complete problem is composed of a maze and an initial state. Although the problem of generating complete problems is interesting, we focus on the task of generating initial states and assume that the maze is provided as input. This way we can directly compare the initial states $\beta$ generates with those designed by human experts on a fixed set of mazes. 
%Namely, we compare the initial states generated by $\beta$ with those of the original puzzles.
%
%We focus on initial states to be able to meaningfully and directly compare the states generated by our system with those designed by human experts. 
%That is, in our evaluation with Sokoban we use the mazes created by humans and directly compare the initial states. Moreover, previous work on Sokoban have also divided the task of generating the maze and the initial state; e.g., \cite{Murase1996,Taylor2011}.
Further, $\beta$ can be used to generate initial states for the problems generated by previous systems. 

The systems of \citeauthor{Murase1996}~\shortcite{Murase1996} and \citeauthor{Taylor2011}~\shortcite{Taylor2011} were used to generate sets of instances to be  
%from which 
employed in the learning process of 
reinforcement learning (RL) algorithms %algorithms can  learn~
\cite{Racaniere2017}. The drawback of using these generative systems is that they are unable to generate large problems, thus limiting RL algorithms to small and easy problems. Since our method generates hard initial states for large Sokoban problems, it will allow RL algorithms to be evaluated in more challenging problems. In the same line of research, \citeauthor{JustesenTBKTR18}~\shortcite{JustesenTBKTR18} showed how systems that generate video game levels can enhance deep RL algorithms. 

%Also related to RL algorithms, 
\citeauthor{Arfaee2011}~\shortcite{Arfaee2011} describe a system that generates a set of initial states of a given problem and learns a heuristic function while solving this set of states with IDA$^*$~\cite{Korf1985}. \citeauthor{Arfaee2011} experimented with domains for which it is easy to generate a set of solvable states with varied difficulty (e.g., sliding-tile and pancake puzzles). Before our work, \citeauthor{Arfaee2011}'s system could not be directly applied to problem domains in which it is hard to generate difficult and solvable states, such as Sokoban. The combination of \citeauthor{Arfaee2011}'s system with $\beta$ is a promising direction of future research. 
%More generally, systems for content generation have been applied 

%Our research falls under the Procedural Content Generation (PCG) umbrella as $\beta$ automatically generates content. 
Procedural Content Generation (PCG) methods are used to generate content such as game levels~\cite{SnodgrassO2017}, puzzles~\cite{SmithM11,SturtevantO18}, and  educational problems~\cite{PolozovOSZGP15}. 
%PCG systems are often specific to a particular game, which contrasts with our approach, that is general to planning problems. 
Many of the problem domains used in the PCG literature (e.g., math problems for children) are computationally easier than the \PSPACE-complete domain we consider in this paper. 
Moreover, in contrast with our work, PCG systems are often concerned with properties such as the enjoyability of the problems generated~\cite{MarinoL16}, as opposed to the scalability of the system. 
Finally, to the best of our knowledge, $\beta$ is the first PCG method employing concepts such as PDB and novelty to the task of problem generation. 

\section{Background and Problem Definition}

Although we focus on the generation of initial states of Sokoban puzzles, our system is general and we describe it as such. 
Let $\mathcal{P} = \tuple{S,s_0,S^*,A}$ be a \emph{state-space search problem}. A state is a complete assignment over a set of variables~$\mathcal{V}$, where each variable $v\in\mathcal{V}$ has a finite set domain $D_v$. Let~$S$ be a set of states, $s_0 \in S$ an initial state, $S^*\subseteq S$ a set of goal states, $A$ a finite set of actions.
An action $a\in A$ is a pair $\tuple{pre_a, post_a}$ specifying the preconditions and postconditions (effect) of action $a$. $pre_a$ and $post_a$ are assignments of values to a subset of variables in $\mathcal{V}$, denoted $\mathcal{V}_{pre}$ and $\mathcal{V}_{post}$. Action $a$ is applicable to state $s$, denoted $s\apply{a}$, if $s$ agrees with $\mathcal{V}_{pre}$. The result of $s\apply{a}$ is state $s'$ that agrees with $post_a$ for the variables in $\mathcal{V}_{post}$ and with $s$ for the remaining variables. We say that the application of $a$ to $s$ \emph{generates} state $s'$. We refer to the states that can be generated from state $s$ as \emph{successor states}, denoted $\successor(s)$. 
We assume that for each action~$a\in A$ there is an \emph{inverse action} $a^{-1} \in A^{-1}$ such that $s' =a\apply{s}$ iff $s =a^{-1}\apply{s'}$. Here, $A^{-1}$ is the set of inverse actions that can applied to $s$, which yield a set of \emph{predecessor states}, denoted $\predecessor(s)$. 
A state $s$ is \emph{expanded} when either $\successor(s)$ or $\predecessor(s)$ is invoked. 

A \emph{solution path} is a sequence of actions that transforms $s_0$ in a state $s_*\in S^*$. 
An \emph{optimal solution path} is a solution path with minimum length. % whose cost is denoted $\mathcal{C^*}$. 
$h(s)$ is a heuristic function that estimates the length to reach a goal state from $s$, with $h^*$ providing the optimal solution length for all states $s\in S$.  A heuristic~$h$ is \emph{admissible} if $h(s)\leq h^*(s)$ for all $s\in S$. 

We can now define the task we are interested in solving. 

\begin{definition}[initial state generation]
An initial state generation task is defined as $\mathcal{P}_{-s_0} = \tuple{S,S^*,A}$. $\mathcal{P}_{-s_0}$ is a state-space search problem $\mathcal{P}$ without a initial state $s_0$. In problem $\mathcal{P}_{-s_0}$ one has to provide a state $s\in S$ such that $P_{-s_0}$ with~$s$ as initial state is solvable. 
\end{definition}

\section{Pattern Databases}

A \emph{pattern database} (PDB) is a memory-based heuristic function defined by a \emph{pattern} $V$ for a problem $\mathcal{P}$~\cite{CulbersonSchaeffer1996,Edelkampecp01}. 
A pattern is a subset of variables $V\in \mathcal{V}$ that induces an abstracted state-space of $\mathcal{P}$. The abstracted space is obtained by ignoring the information of variables $v\notin V$. For a given pattern, the optimal solutions of all states in the induced abstracted space is computed and stored in a look-up table. These stored values can then be used as a heuristic function estimating $h^*$. The combination of multiple PDB heuristics can result in better estimates of $h^*$ than the estimates provided by each PDB alone. PDBs can be combined by taking the maximum~\cite{holte2006a} or the sum~\cite{felner:etal:jair-04} of multiple PDB estimates. The heuristic function resulting from the sum of multiple PDBs is called an additive PDB heuristic. An additive PDB is admissible if no action affects more than one pattern $V$, where an action $a$ affects $V$ if $a$ has an effect on any of its variables, i.e., $\mathcal{V}_{post_a} \cap V \neq \emptyset$. Consider $\mathcal{P}_e$, the search problem shown in Figure~\ref{fig:example}, and the example that follows. 

\begin{figure}[h]
  \begin{tabular}{rcl}    
    $\mathcal{V}$ & $=$ & $\{v_1,v_2,v_3 \}$,\\
    $D_{v_i}$ & $=$ & $\{0,1,2,3,4\}$ for all $v_i \in \mathcal{V}$, \\
    $s_0$ & $=$ & $\{v_1=0,v_2=0,v_3=0 \}$,\\
    $s_*$ & $=$ & $\{v_1=3,v_2=3,v_3=3 \}$,\\
    $A$ & $=$ & $\{inc^v_x | v\in\mathcal{V}, x\in\{0,1,2\}\}$ $\cup$ $\{jump^v | v\in\mathcal{V}\}$,  \\
    \multicolumn{3}{c}{$inc^{v_i}_x = \langle v_i=x, v_i=x+1 \rangle$,} \\
    \multicolumn{3}{c}{$jump_{v_i} = \langle \mathcal{V}-v_i=4 \land v_i=0, v_i=3\rangle$,} \\
%    \multicolumn{3}{c}{with $\cost_a=1$ for all $a \in A$.} \\
  \end{tabular}
  \caption{Example 
  %of a state-space search problem 
  adapted from~\protect\citeauthor{pommerening:etal:ijcai-13}~\protect\shortcite{pommerening:etal:ijcai-13}.}
  \label{fig:example}
  \end{figure}
  
$\mathcal{P}_e$ has three variables that can be assigned a value in $\{0, 1, 2, 3, 4 \}$. The initial state assigns $0$ to all variables, and the goal state has all variables with the value of $3$. $\mathcal{P}_e$ has two types of actions $inc^{v_i}_x$ and $jump_{v_i}$. Actions $inc^{v_i}_x$ change the value of the variable $v_i$ from $x$ to $x+1$. Actions $jump_{v_i}$ change the value of variable $v_i$ from $0$ to $3$, as long as all the other variables have the value of $4$. All solutions to this problem involve the use of action $inc^{v_i}_x$ three times to all three variables, yielding an optimal solution length of $9$. $jump_{v_i}$ cannot be used in any solution because if a variable has the value of $4$, the problem becomes unsolvable since there is no action to change its value.

\begin{example}
  Let $\{\{v_1\}, \{v_2, v_3\}\}$ be a collection with two additive patterns for $\mathcal{P}_e$. The heuristic estimate returned for $\mathcal{P}_e$'s $s_0$ with a PDB heuristic with pattern $\{v_1\}$ is $1$, as a single action $jump$ changes the value of $v_1$ from $1$ to $3$, thus achieving the goal in the PDB's abstracted space.
 % Note that the action $jump$ is applicable because the other two variables are ignored in the pattern $\{v_1\}$.
The heuristic estimate returned for $s_0$ for pattern $\{v_2, v_3\}$ is $6$. This is because the action $jump$ is no longer applicable.
% , and the solution to the abstracted problem induced by the pattern is to apply the action $inc$ three times for each variable.
Since the two patterns are additive, an additive PDB with $\{\{v_1\}, \{v_2, v_3\}\}$ returns $1 + 6 = 7$ as an estimate of the solution cost of $s_0$.
\label{example_pdb}
\end{example}

\section{Proposed Search Algorithm for $\mathbf{\beta}$}

%In this section we present our proposed search procedure $\beta$.

%\subsection{Search Procedure for Initial State Generation}

Given a problem $\mathcal{P}_{-s_0}$, $\beta$ performs a backward greedy best-first search~\cite{Doran1966} from the set $S^*$. 
This is performed by inserting all states $s_*\in S^*$ in a priority queue denoted \texttt{open} and in a set of generated states denoted \texttt{closed}. 
$\beta$ expands states from \texttt{open} according to an order $[\cdot]$, which $\beta$ receives as input.
When a state $s$ is removed from \texttt{open}, $\beta$ expands $s$ invoking $\predecessor(s)$. Every $s'\in \predecessor(s)$ that is not in \texttt{closed} is inserted in \texttt{open} and \texttt{closed}. 

The advantage of performing a backward search is that all states generated that way are guaranteed to have a solution. By contrast, if one assigned random values from the variables' domains as a means of generating initial states, then one would need to prove the resulting $\mathcal{P}$ problem to be solvable, and in some domains, this task is computationally hard. 
$\beta$ returns the state $s$ encountered in search with largest \emph{objective value} once a time or a memory limit is reached. We specify below the objective functions used with $\beta$. 

%subsection{Proposed Metrics of Difficulty}
\subsection{Solution Length}
We propose two metrics of difficulty: solution length and conflicts.
%\subsubsection{Solution Length}
The optimal solution length is traditionally used to evaluate the hardness of initial states for AI methods.
For example, solution length correlates with the running time of heuristic search algorithms in problems such as Rubik's Cube~\cite{LelisZH2013b}. 
We use PDB heuristics to approximate the optimal solution length of state-space problems. 
%Since we are interested in the optimal solution length, in domains with explicit actions costs we simple ignore the costs.

\subsection{Conflicts}

Intuitively, the hardness of state-space problems arises from the interactions among variables. If sub-problems defined by single variables of the problem can be solved independently, then the problem tends to be easy. However, if one needs to consider interactions between all variables of the problem, then the problem tends to be hard.
We use PDBs to capture the interaction of variables in a problem. Let $h^{\text{PDB}_k}$ be a PDB heuristic with patterns of size $k$. The \emph{conflicts} of state $s$ is defined as follows. 

\begin{definition}[conflicts]
  The number~$n$ of \emph{conflicts} of order~$k$, denoted $kC$, for $k>1$ of a state~$s$ is $n=h^{\text{PDB}_k}(s)-h^{\text{PDB}_{k-1}}(s)$, if $n>0$; $n = 0$, otherwise. 
  %if $h^{\text{PDB}_k}(s)-h^{\text{PDB}_{k-1}}(s) < 0$.
\end{definition}

Suppose we have the following PDB heuristics for our running example: $h^{\text{PDB1}}$, $h^{\text{PDB2}}$ and $h^{\text{PDB3}}$. $h^{\text{PDB1}}$ uses three additive patterns $\{\{v_1\},\{v_2\},\{v_3\}\}$, which results in $h^{\text{PDB1}}(s_0) = 3$ (the action $jump$ is applicable to each pattern). 
$h^{\text{PDB2}}$ uses two additive patterns, one with two variables and another with one variable: $\{\{v_1\},\{v_2,v_3,\}\}$. For these patterns $h^{\text{PDB2}}(s_0) = 7$. Finally, $h^{\text{PDB3}}$ uses one pattern with all three variables and returns the optimal solution of $9$. 

Using these PDB-values we can compute the conflicts of $s_0$. $s_0$  has four conflicts of order two, denoted $2C$ ($h^{\text{PDB2}}(s_0) - h^{\text{PDB1}}(s_0) = 4$) and two conflicts of order three, denoted $3C$ ($h^{\text{PDB3}}(s_0) - h^{\text{PDB2}}(s_0) = 2$). The number of $2C$ conflicts approximates the effort in terms of solution length due to the interaction of pairs of variables. In general, the number of $kC$ conflicts approximates the effort due to the interaction of $k$ variables. %We conjecture that many conflicts of large $k$ result in the harder initial states.

\subsection{Novelty}

Next, we discuss novelty, which we use to enhance the exploration of our search algorithm.
Novelty~$w$ is a structural measure proposed to enhance the exploration of search algorithms in the task of finding solutions for search problems~\cite{Lipovetzky2012,Lipovetzky2017}.
Novelty evaluates states with respect to the order in which an algorithm generates them. 

\begin{definition}[novelty]
Let $\mathcal{P}$ be a state-space search problem, $s \in S$ be a state generated by a search algorithm, and $h$ be a heuristic function. The \emph{novelty} of~$s$ for~$h$ is $|\mathcal{V}| - n + 1$ if $s$ is the first generated state during search with heuristic value $h(s)$ and with a subset of~$n$ variables being assigned values $d_{v_1}, \cdots, d_{v_n}$ and no smaller subset of variables has this property. 
\label{novelty-h}
\end{definition}

The novelty of a state $s$ generated during the search is maximal if, for the first time in search, $s$ has a variable $v$ for which the value $d_v\in D_v$ is assigned. The novelty of $s$ is minimum if $n = |\mathcal{V}|$, which means that it requires all variables in $\mathcal{V}$ to differ $s$ from previously generated states. Intuitively, a state has larger novelty if it requires a smaller number of variables to assume a previously unseen assignment of values.\footnote{Our semantics for novelty differs from \citeauthor{Lipovetzky2012}'s semantics because they dealt with minimization problems while we deal with a maximization problem.}

\begin{example}
Suppose we run $\beta$ in $\mathcal{P}_e$ with the ordering $[\cdot]$ defined by novelty and that all states have the same heuristic value. State~$s_*$ has novelty~$w(s_*)=3$ because this is the first state to assign $3$ to  $v_1$. Then, $\beta$ invokes $\predecessor(s_*)$ and generates three states $s_1, s_2, s_3$ by applying the $inc^{-1}$ (the inverse of $inc$) actions to each variable. For simplicity, we will assume that $jump^{-1}$ is not applicable. $s_1, s_2, s_3$ have $w=3$ because each of them assigns $2$ to a variable for the first time. Next, suppose $\beta$ expands $s_1=\{v_1=2,v_2=3,v_3=3 \}$, then it generates three states $\{v_1=1,v_2=3,v_3=3 \}$, $\{v_1=2,v_2=2,v_3=3 \}$ and $\{v_1=2,v_2=3,v_3=2 \}$, where the last two states have $w=2$ because the smallest subset with newly assigned values is two: $\{v_1=2,v_2=2\}$ for the second and $\{v_1=2,v_3=2\}$ for the third. % Finally, suppose that  $\beta$ generates $s_0$  as the last state before generating the whole state-space, then $s_0$  has $w=1$ as the assignment of $0$ to all three variables in $s_0$ is the smallest subset not observed until then.
\end{example}

We hypothesize that novelty can  improve the exploration of the search performed by $\beta$, thus allowing it to visit a diverse set of states from which one can select initial states that maximize solution length and number of conflicts. 

\section{Difficulty Metrics for Humans in Sokoban}

\citeauthor{Jaruvsek2010}~\shortcite{Jaruvsek2010} performed an extensive user study showing strong positive correlations between metrics of difficulty they introduced with the time required by humans to solve Sokoban problems. In this section we introduce the domain of Sokoban and  present their metrics. %We also argue that \citeauthor{Jaruvsek2010}'s metrics are closely related to our PDB-based metrics.

\subsection{Sokoban}

Sokoban is a transportation problem played in a maze grid. The maze is defined by locations occupied by immovable blocks (walls), free (inner) locations and goal (inner) locations. 
An initial state of Sokoban is a placement of $k$ boxes and a man in inner locations of the maze.
%A set of $k$ boxes and a man are placed in free locations. 
The goal is to push, through the man, all boxes from their initial location to a goal location while minimizing the number of pushes.
%The map is defined by walls, inner and goal locations. An initial state is a placement of the $k$ boxes and the man.

\begin{figure}[t]
    \centering
    \includegraphics{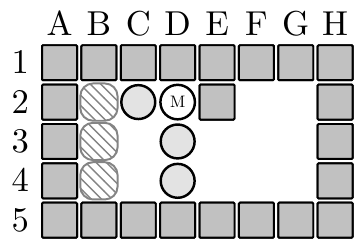}
    \caption{Sokoban problem.}
    \label{fig:2c}
\end{figure}

Figure~\ref{fig:2c} shows a Sokoban maze with a wall at A$1$, an free location at C$3$ and goal location at B$4$. 
An initial state is defined by the assignment of the locations of boxes and the man. A state has a $k+1$ variables, one for each box and another for the man. The instance has $k$ goal locations, one for each box. The domain of each variable is the set of inner locations. 
Goal states have all~$k$ boxes at one of the~$k$ goal locations.
%\begin{wrapfigure}{l}{0.2\textwidth}
%         \resizebox{1.\linewidth}{!}{\includegraphics[page=1]{Figures/Figures} }        
% \caption{Sokoban problem.}
%     \label{fig:2c}
% %\end{wrapfigure}
Figure~\ref{fig:2c} shows a box at C$2$ and the man at D$2$. 
We name goal locations and boxes by their locations in the maze. For example, the box at C$2$ is box C$2$.
An action corresponds to the man pushing a box to an adjacent empty inner location. The precondition of an action is that the man must be able to reach the orthogonal location adjacent to the box. For example, in Figure~\ref{fig:2c} box C$2$ can be pushed to B$2$, but box D$3$ cannot be pushed anywhere. 

\subsection{\protect\citeauthor{Jaruvsek2010}'s Metrics}

The first metric introduced by \citeauthor{Jaruvsek2010}~\shortcite{Jaruvsek2010} was solution length, which is exactly what we approximate with PDBs. \citeauthor{Jaruvsek2010} showed that the exact solution length yields a Spearman correlation of $0.47$ with the time required by humans to solve Sokoban problems. 

\citeauthor{Jaruvsek2010}~\shortcite{Jaruvsek2010} introduced a metric named ``number of counterintuitive moves''. In Figure~\ref{fig:2c}, while it is intuitive to push box C$2$ to a goal location (push it to the left), the pushes required to move boxes D$3$ and D$4$ to a goal location are ``counterintuitive''. This is because one needs to push either D$3$ or D$4$ away from a goal location before pushing them to a goal location. These counterintuitive pushes can be approximated by our metric of conflicts. In Figure~\ref{fig:2c}, a PDB that considers only one box estimates that one can directly push all three boxes to the goal, yielding an estimate of $1 + 2 + 2 = 5$ pushes: one for C$2$ and two for D$3$ and D$4$. A PDB that considers two boxes captures the interaction between D$3$ and D$4$. Such a PDB discovers that one needs to push either D$3$ or D$4$ two locations to the right, before pushing them to a goal location, for an estimated solution length of $1 + 6 + 2 = 9$. The difference $9 - 5 = 4$ is the number of \citeauthor{Jaruvsek2010}'s counterintuitive moves captured by our metric of conflicts. \citeauthor{Jaruvsek2010} showed that the exact number of conflicts has a Spearman correlation of $0.69$ with the time required by humans to solve Sokoban problems. 

 \citeauthor{Jaruvsek2010}~\shortcite{Jaruvsek2010} also introduced ``problem decomposition'', a metric that measures the degree in which subsets of boxes can be independently pushed to the goal.
This metric yields a Spearman correlation of 0.82 with the time required by humans to solve the puzzles. 
Our order $k$ of conflicts approximates similar property. If $k$ is the largest value for which $h^{\text{PDB}_{k}}(s)-h^{\text{PDB}_{k - 1}}(s) > 0$ for all possible PDBs with $k$ and $k-1$ variables, then the largest number of variables with counterintuitive moves in $s$ is $k$. Although one might be unable to divide $s$ into independent subproblems with $k$ variables (e.g., the order in which the subproblems are solved might matter), one is able to reason about the subproblems with at most $k$ variables somewhat independently.

The metrics introduced by \citeauthor{Jaruvsek2010}~\shortcite{Jaruvsek2010} are computationally hard in general as they require one to solve Sokoban problems to compute their values. 
By contrast, our metrics can be computed efficiently and thus be used within search procedures for generating hard initial states. 

\section{Empirical Evaluation}

\begin{table*}[h]
\scalefont{0.85}
%\resizebox{\textwidth}{!}{%\tiny
\centering
\begin{tabular}{lllrrrrrrrrrr}
\toprule
& \multirow{2}{*}{\#} & \multirow{2}{*}{Generation Method} & \multirow{2}{*}{\begin{tabular}[c]{@{}c@{}}States\\ Expanded\end{tabular}} & \multicolumn{4}{c}{$h$-value of $s_0$} & \multicolumn{3}{c}{Number of Conflicts} & \multicolumn{1}{c}{\multirow{2}{*}{\# Solved}} \\ 
\cmidrule(lr){5-8}\cmidrule(lr){9-11} 
&  & & &
\multicolumn{1}{c}{$h^{\text{PDB1}}$} &
\multicolumn{1}{c}{$h^{\text{PDB2}}$} &
\multicolumn{1}{c}{$h^{\text{PDB3}}$} &
\multicolumn{1}{c}{$h^{\text{PDB4}}$} & 
\multicolumn{1}{c}{2C} & 
\multicolumn{1}{c}{3C} & 
\multicolumn{1}{c}{4C} & 
 \\ % \cite{Pereira2016b}} \\ 
\midrule
\multirow{4}{*}{A} & 1 & $[ h^{\text{PDB1}}]$ & 36,844,914.09 & 179.68 & 185.94 & 187.88 & 190.49 & 6.27 & 1.93 & 2.61 & 33.0 $\pm$ 0.0 \\
 & 2 & $[ h^{\text{PDB2}}]$ & 35,746,724.37 & 178.70 & 190.33 & 190.45 & 193.35 & 11.63 & 0.11 & 2.90 & 35.0 $\pm$ 0.0 \\
 & 3 & $[ h^{\text{PDB3}}]$ & 16,762,063.67 & 166.28 & 175.87 & 178.09 & 180.63 & 9.58 & 2.22 & 2.54 & 33.8 $\pm$ 1.8 \\
 & 4 & $[ h^{\text{PDB4}}]$ & 19,683,953.17 & 168.01 & 177.78 & 179.26 & 183.63 & 9.77 & 1.48 & 4.37 & 32.4  $\pm$ 2.8 \\
\midrule
\multirow{3}{*}{B} & 5 & $[h^{\text{PDB2}},2C]$ & 35,279,521.51 & 179.08 & 191.31 & 191.32 & 194.31 & 12.23 & 0.01 & 2.99 & 36.0  $\pm$ 0.0\\
 & 6 & $[h^{\text{PDB3}},3C]$ & 14,896,276.53 & 167.41 & 176.91 & 179.70 & 182.31 & 9.49 & 2.79 & 2.62 & 33.2 $\pm$ 1.9 \\
 & 7 & $[h^{\text{PDB4}},4C]$ & 10,127,736.49 & 157.09 & 166.24 & 167.84 & 172.41 & 9.15 & 1.60 & 4.57 & 33.0 $\pm$ 2.5 \\
\midrule
\multirow{3}{*}{C} & 8 & $[2C, h^{\text{PDB2}}]$ & 23,465,651.94 & 136.89 & 158.97 & 154.48 & 158.78 & 22.08 & 0 & 4.30 & 26.0 $\pm$ 0.0 \\
 & 9 & $[3C, h^{\text{PDB3}}]$ & 10,075,985.22 & 111.44 & 117.76 & 126.85 & 127.69 & 6.31 & 9.10 & 0.83 & 28.8 $\pm$ 0.8 \\
 & 10 & $[4C, h^{\text{PDB4}}]$ & 6,066,057.87 & 101.95 & 111.48 & 112.42 & 123.57 & 9.54 & 0.94 & 11.15 & 31.0 $\pm$ 4.9 \\
\midrule
\multirow{4}{*}{D} & 11 & $[w(h^{\text{PDB1}}), h^{\text{PDB1}}]$ & 24,852,925.84 & \textbf{223.43} & 231.11 & 233.79 & 237.11 & 7.68 & 2.68 & 3.32 & 26.6 $\pm$ 0.5 \\
 & 12 & $[w(h^{\text{PDB2}}), h^{\text{PDB2}}]$ & 23,764,256.42 & 219.52 & \textbf{233.19} & \textbf{234.14} & \textbf{238.07} & 13.67 & 0.96 & 3.92 & 29.8 $\pm$ 0.4 \\
 & 13 & $[w(h^{\text{PDB3}}), h^{\text{PDB3}}]$ & 12,320,342.90 & 215.49 & 226.89 & 230.16 & 233.44 & 11.40 & 3.27 & 3.28 & 25.4 $\pm$ 1.1 \\
 & 14 & $[w(h^{\text{PDB4}}), h^{\text{PDB4}}]$ & 12,643,046.69 & 190.76 & 202.03 & 204.12 & 210.70 & 11.27 & 2.08 & 6.59 & 25.8 $\pm$ 1.5 \\
\midrule
\multirow{3}{*}{E} & 15 & $[w(h^{\text{PDB2}}), h^{\text{PDB2}}, 2C]$ & 26,994,420.86 & 216.29 & 230.81 & 231.22 & 235.19 & 14.52 & 0.40 & 3.97 & 27.0 $\pm$ 0.0 \\
 & 16 & $[w(h^{\text{PDB3}}), h^{\text{PDB3}}, 3C]$ & 11,842,771.67 & 211.28 & 222.30 & 225.45 & 228.89 & 11.02 & 3.15 & 3.45 & 26.0 $\pm$ 1.2 \\
 & 17 & $[w(h^{\text{PDB4}}), h^{\text{PDB4}}, 4C]$ & 7,650,277.40 & 207.38 & 218.77 & 221.24 & 227.21 & 11.39 & 2.47 & 5.97  & 26.2 $\pm$ 2.9 \\
\midrule
\multirow{3}{*}{F} & 18 & $[w(h^{\text{PDB2}}), 2C, h^{\text{PDB2}}]$ & 15,280,827.99 & 175.46 & 201.34 & 196.69 & 202.62 & \textbf{25.89} & 0 & 5.93 & 22.8 $\pm$ 0.4 \\
 & 19 & $[w(h^{\text{PDB3}}), 3C, h^{\text{PDB3}}]$ & 7,412,556.34 & 154.76 & 162.73 & 175.78 & 175.85 & 7.96 & \textbf{13.06} & 0.07 & 20.6 $\pm$ 2.1 \\
 & 20 & $[w(h^{\text{PDB4}}), 4C, h^{\text{PDB4}}]$ & 
   3,986,192.80 &  148.37 	&  159.96 & 161.09 & 176.33 & 11.60 & 1.14 & 15.24 & 19.5 $\pm$ 2.4 \\
 %4,632,128.63 & 148.35 & 160.03 & 161.23 & 176.55 & 11.68 & 1.20 & 15.32 & 18.1 $\pm$ 3.0\\
 \midrule
 %$[w(h^{\text{PDB4}}), 4C, h^{\text{PDB4}}]^*$
 %& 21 & Aggregation & 46,321,286.67  &  149.24 & 160.49 & 159.78 & 183.41 & 11.27 & 0.82 & \textbf{23.63} & \textbf{15}\\ 
 & 21 & Aggregation &
   79,699,027.84  & 151.46 & 163.07 & 161.78 & 187.00 & 11.63 & 0.77 & \textbf{25.22} & 
 %44,141,938.58 & 152.61 & 164.21 & 163.29 & 187.12 & 11.39 & 0.85 & \textbf{23.83} & 
 \textbf{16.4 $\pm$ 1.3}\\
 \midrule
  %& 22 & Random Walk & 77,942,578.73 & 12.42 & 14.16 & 14.07 & 14.45 & 1.74 & 0.12 & 0.38 & 86.5 $\pm$ 1.8 \\
  %& 23 & Breadth First Search & 29,057,981.78 & 38.90 & 41.27 & 41.65 & 42.42 & 2.37 & 0.38 & 0.77 & 85.2 $\pm$ 0.5\\ 
 %\midrule
-  & & Humans (xSokoban) & -  & 188.02 & 199.58 & 205.62 & 211.41 & 11.56 & 6.04 & 5.79  & 18 \\ \bottomrule
\end{tabular}%}
\caption{Results of states generated by variants of $\beta$. The ``$h$-value of $s_0$'' denotes the average heuristic value of the generated initial states; ``Number of Conflicts'' shows the average number of conflicts, where $2$C, $3$C, and $4$C denote the number of conflicts of order $2$, $3$, and $4$, respectively. ``\# Solved'' denotes the average and standard deviation of the number of problems solved by PRB. 
%The last line shows the same information for initial states created by human experts.
} \label{tab:results}
\end{table*}

%\paragraph{Problem Domain.} 
%\vspace{0.1in}

%\textbf{Problem Domain} %We evaluate $\beta$  
%guided by our hardness metrics and novelty 
%on the problem domain of Sokoban. 
We use the $90$ problems of the xSokoban benchmark in our experiments. These problems were developed by human experts and are known to be challenging to both humans and AI methods. The xSokoban benchmark allows us to directly compare the initial states our method generates with those created by human experts in terms of the metrics proposed and number of problems solved by a solver. Moreover, the $\mathcal{P}_{-s_0}$ problems present in xSokoban contain many more boxes (the largest problem has $34$ boxes) than the problems generated by the current generative methods (the largest problem reported in the literature has $7$ boxes). 

%\textbf{Sokoban Solver}
%We hypothesize that Sokoban initial states with large estimated solution lengths and large number of conflicts of high order are hard for a state-of-the-art solver. The solver we use is the one introduced by \citeauthor{Pereira2015}~\shortcite{Pereira2015}, which we will refer to as PRB.  
%PRB is based on \astar{}~\cite{Hart1968} with a domain-dependent tie-breaking strategy and a sophisticated heuristic function.

%\textbf{Methods Evaluated} 
We instantiate variants of $\beta$ by varying the ordering $[\cdot]$ in which the states are expanded and how initial states are selected. $[\cdot]$ is composed by a list of features, which can include novelty $w(h)$, a PDB heuristic $h^{\text{PDBk}}$, and the number of conflicts of order $k$, denoted $kC$. $\beta$ returns the state generated with largest ordering value, ignoring novelty, once the search reaches the time or memory limit. For example, $\beta$ with ordering $[w(h^{\text{PDB4}}), 4C, h^{\text{PDB4}}]$ expands states with largest $w(h^{\text{PDB4}})$ first, with the ties broken according to $4$C and then according to $h^{\text{PDB4}}$. $\beta$ returns the state $s$ with largest $4$C-value, with ties broken according to $h^{\text{PDB4}}$; remaining ties are broken by the state generation order. % in which the states are generated.

%\textbf{PDBs} 
Heuristic $h^{\text{PDBk'}}$ returns the maximum heuristic value for a set with $|\mathcal{V}|+1$ additive PDBs. We use this number of PDBs to ensure that instances with more variables have more PDBs. The patterns of each of the $|\mathcal{V}| + 1$ additive PDBs are defined by iteratively and randomly selecting without replacement $k'$ variables from $\mathcal{V}$, until all variables are in a pattern. If $|\mathcal{V}| \mod k' > 0$, then 
%by selecting variables without replacement one will eventually be left with $|\mathcal{V}| \mod k'$ variables. In this case, 
the last pattern will have $|\mathcal{V}| \mod k'$ variables. 
%The remaining additive PDB sorts the state according the value of each variable and builds each pattern linearly. 
All patterns include the man.

 %\textbf{Solver} 
 %We hypothesize that Sokoban initial states with large estimated solution lengths and large number of conflicts of high order are hard for a state-of-the-art solver. 
 The solver we use to evaluate the hardness of the instances generated by $\beta$ and by human experts is the one introduced by \citeauthor{Pereira2015}~\shortcite{Pereira2015}, which we refer to as PRB.

%\textbf{Metrics} 
We are primarily interested in comparing the states generated $\beta$ with those designed by human experts in terms of problems solved by PRB. 
%Although our main metric is the number of problems solved by the PRB solver, 
%We also present metrics used to explain the number of problems solved by PRB. 
We also present the average number of states expanded by each variant of $\beta$ while generating the 90 initial states, the average $h$-value of the initial states generated in terms of $h^{\text{PDB1}}$, $h^{\text{PDB2}}$, $h^{\text{PDB3}}$ and $h^{\text{PDB4}}$, and the average number of conflicts 2C, 3C, and 4C. We also experimented with random walks (RW) and breadth-first search (BFS) as the search algorithms used by $\beta$ and PRB solves almost all problems generated by these approaches (approximately 85 out of 90). This is because both methods generate initial states with very short solution length. RW and BFS results are omitted from the table of results for clarity.

%\textbf{Experimental Setting} 
All experiments are run on $2.66$ GHz CPUs, $\beta$ is allowed 1 hour of computation time and $8$ GB of memory,  
while the solver is allowed $10$ minutes and $4$ GB (more memory and running time have a small impact on PRB). 
%result in major changes to the results.
Due to the 
%stochastic nature of our PDBs, 
randomness of the PDBs, 
we 
%repeat each experiment $5$ times and 
report the average results of $5$ runs of each approach. We also report the standard deviation for the number of problems PRB solves. 
Table~\ref{tab:results} presents the results of our proposed method. The third column of the table (``Generation Method'') shows the order used to guide the $\beta$ search and to select initial states in our experiments. We also present the number of states $\beta$ expanded, number of problems solved by PRB, and the $h$-values and conflict values, the table also presents a column ``\#'' indicating the row of each method, which are used as a reference in our discussion. 
We highlight in bold of cells with the best result for a given criterion. For example, $[w(h^{\text{PDB1}}), h^{\text{PDB1}}]$ is the best performing method in terms of value of  $h^{\text{PDB1}}$. 

\subsection{Quantitative Results}

%\textbf{Solution Length} 
We start by comparing the $\beta$ results without novelty (see blocks $A$, $B$ and $C$ in Table~\ref{tab:results}) with those with novelty (blocks $D$, $E$ and $F$). 
%As we have hypothesized, 
%$\beta$ guided by novelty can generate a more diverse set of states. 
Methods that use novelty outperform their counterparts in all criteria: average $h$-value, number of conflicts, and the total number of problems solved. Compare, for example, the results in rows $1$--$4$ with their counterparts in rows $11$--$14$. In particular, $[ w(h^{\text{PDB1}}), h^{\text{PDB1}}]$ obtained an average $h^{\text{PDB1}}$-value of $223.43$, which is larger than the value of $188.02$ for the average of the xSokoban initial states. 
Although $[w(h^{\text{PDB1}}), h^{\text{PDB1}}]$ can generate initial states with larger $h^{\text{PDB1}}$ values than the xSokoban, its initial states tend to be much easier in terms of the number of problems solved than the xSokoban. PRB solves at least seven more initial states generated by the methods in block $D$ than the initial states created by human experts. 
%This is likely because the initial states generated by the methods in block $D$ have fewer conflicts of higher order (e.g., $4$C-value of $5.26$ for row 14, while xSokoban has a 4C-value of 5.79). 

%\textbf{Conflicts} 
The methods shown in blocks $B$ and $E$ maximize the heuristic value and break ties by conflict values. Methods in blocks $C$ and $F$ use the opposite order of metrics. The methods in $B$ generate initial states with longer solution lengths than the methods in $C$. However, the methods in $C$ have a larger number of conflicts, according to the optimized criteria. For example, the method in row $8$, which prioritizes the maximization of $2$C-values, generates initial states with larger $2$C-values than its counterpart in row $5$. In general, the initial states generated by the methods in $C$ tend to be much harder in terms of the number of problems solved by PRB. These results demonstrate the importance of generating initial states with a large number of conflicts. The same pattern is observed by comparing the results of the methods in $E$ with those of the methods in $F$. The results in $F$ highlight the importance of combining the maximization of conflicts of higher order with novelty to enhance exploration. The method shown in line 20 is already competitive with the xSokoban benchmark in terms of number of problems solved by PRB.
%of high order to generate hard problems, as the average number of problems solved decreases as the number of conflicts of high order increases. 
%The hardest initial states we are able to generate are those in row $20$, which maximizes conflicts of order $4$. For this method, PRB is able to solve only 18.1 problems on average.  
%The method in row $20$ generates initial states with smaller $h$-values, but much larger number of conflicts of high order, which resulted in harder initial states. Thus, in the context of generating hard initial states for an AI-based solver, our $\beta$ variant presented in row $20$ is able to outperform the initial states designed by human experts. 

%Memory is one of the main limitations of $\beta$ search, as the number of states visited depends on the amount of memory available. One way of reducing the impact 

%\textbf{Aggregation procedure} 

The method presented in line $21$ shows the average numbers of $5$ runs of an aggregation procedure. One run of this procedure represents $20$ runs of $[w(h^{\text{PDB4}}), 4C, h^{\text{PDB4}}]$. 
%of the method presented in line $20$. 
%That is, we perform $20$ independent runs of $\beta$ with the configuration shown in line $20$. 
Then, out of the $20$ runs, we select for each maze the initial state that maximizes $[4C, h^{\text{PDB4}}]$. This procedure further optimizes the average 4C-value (25.22) at the cost of a larger number of state expansions and it   
%(i.e., the average of the sum of all runs). 
%The average 4C-value can only increase as one increases the number of runs, with the resulting instances being more likely to be harder for the solver. 
%The aggregation procedure 
outperforms the initial states designed by humans in terms of number of problems PRB is able to solve. PRB solves 18 xSokoban instances and only an average of 16.4 instances of the aggregation procedure. % with an average of 16.4 instances of the aggregation method being solved by PRB. % with $20$ runs. 
%If using only the first $5$ of such runs, 18 instances are solved (result not shown in the table). 
 
%This result is important because it suggests that $\beta$ could be used to automate part of crafting process of novel problems for evaluating planning systems. % could be automated by a search method based on our work.

%\textbf{Difficulty for Humans} 
%We argued that $\beta$'s metrics are related to the metrics shown to correlate with the time required by humans to solve Sokoban problems~\cite{Jaruvsek2010}. 

The metric that correlated to most with the time required by humans to solve Sokoban problems in \citeauthor{Jaruvsek2010}'s study was ``problem decomposition,'' which is similar to our order of conflict. 
%The results shown in Table~\ref{tab:results} 
Our results 
suggest that states with many conflicts of high order tend to be harder to solve by PRB. 
%That is, the initial states with higher $4$C-values tend to be harder to solve by the PRB solver. 
Taken together with \citeauthor{Jaruvsek2010}'s user study, our results suggest that some of the properties that make problems hard for humans can also make them hard for AI systems. 

\begin{figure}[t]
\centering
\begin{subfigure}[t]{0.45\linewidth}
  \makebox[0pt]{\rotatebox[origin=lt]{90}{Our Approach}\hspace*{2em}}%
  \centering
  \resizebox{1.\linewidth}{!}{\includegraphics[page=2]{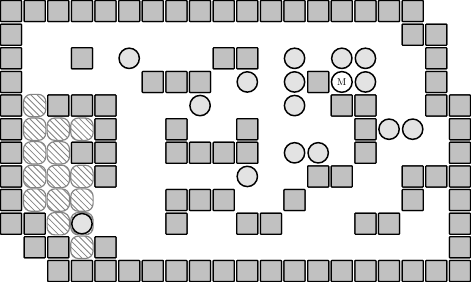} }
  \label{fig:instance_53}
\end{subfigure}
\begin{subfigure}[t]{0.45\linewidth}
  \centering
  \resizebox{1.\linewidth}{!}{\includegraphics[page=3]{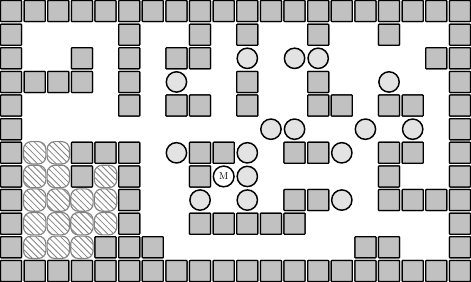} }
  \label{fig:instance_62}
\end{subfigure}

\begin{subfigure}[t]{0.45\linewidth}
\makebox[0pt]{\rotatebox[origin=lt]{90}{
   xSokoban}\hspace*{2em}}%
  \centering
  \resizebox{1.\linewidth}{!}{\includegraphics[page=4]{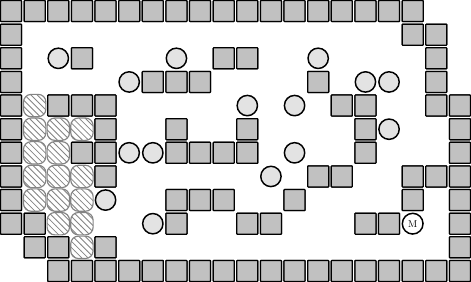} }
    \caption{\#53.}
    \label{fig:sokoban_67}
\end{subfigure}
\begin{subfigure}[t]{0.45\linewidth}
  \centering
  \resizebox{1.\linewidth}{!}{\includegraphics[page=5]{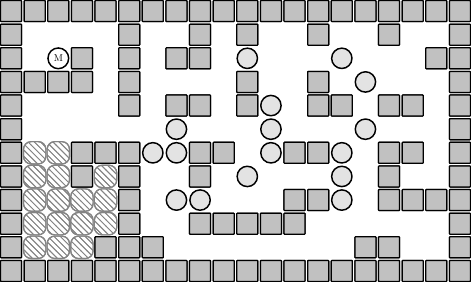} }
    \caption{\#62.}
    \label{fig:sokoban_67}
\end{subfigure}

\caption{Initial states of Sokoban generated by our best variant of $\beta$ (top row), and the initial states of xSokoban (bottom row).}
\label{fig:ai_human}
\end{figure}

%\vspace{0.1in}
%\noindent
\subsection{Qualitative Results}

Figure~\ref{fig:ai_human} shows representative initial states for two xSokoban mazes. The initial states shown at the top were generated by a representative run of $[w(h^{\text{PDB4}}), 4C, h^{\text{PDB4}}]$ and at the bottom are the original xSokoban problems. These initial states are visually similar since there are no obvious features that distinguish those states generated by $\beta$ from those created by humans. 
%The states generated by our approach and those designed by human experts have $1$ and $6$ $4$C conflicts, for mazes \#53 and \#62, respectively.
The states generated by our method have larger values of $h^{\text{PDB4}}$ for these mazes: $219$ and $223$, while the values are $167$ and $202$ for xSokoban. PRB solves the xSokoban problem for maze \#53, but does not solve the $\beta$ state for the same maze. Although both initial states have the same $4$C-value for maze \#53 ($=2$), our initial state has a much longer solution length, which could explain the result. The situation is reversed for maze \#62, where PRB solves the initial state generated by our approach, but does not solve the initial state designed by humans, and both states also have the same $4$C-value ($=6$). We conjecture that the xSokoban state for maze \#62 has conflicts of order higher than $4$, which our PDBs are unable to capture. These conflicts of higher order could make the xSokoban problem harder to solve than the one generated by our system.  
%The initial state of maze \#67 shows another characteristic of initial states generated by our methods, which is to have more boxes on goal locations. On average xSokoban has 0.37 boxes on goal locations while our method in row $20$ has 1.61.

%\section{Discussion}
%\textbf{Discussion}
Currently, no solver is able to solve all xSokoban problems. Yet, humans can solve all of them. Thus, humans are still superior to AI methods in the task of solving Sokoban problems. By contrast, our results show that $\beta$ achieves superhuman performance in the task of generating initial states in terms of instance difficulty for a solver. 
%An interesting direction of future work is to verify if the start states generated by $\beta$ are also harder to humans than those from the xSokoban benchmark. 

%The main limitation of our approach is that the experiments are restricted to Sokoban. However, Sokoban is an important domain by itself and traditional testbed for AI methods. It is one of the hardest search problems due to large branching factor, long solution length, and the presence of unsolvable states. In addition, instances of Sokoban generated by humans are hard. No specialized solver can solve all instances of the xSokoban benchmark. Yet, humans can solve all of them. Thus, humans are superior to AI methods in the task solving instances. Our results showed how to achieve superhuman performance in the task generating initial states in this domain. 

Although we evaluated $\beta$ only on Sokoban, we believe it can generalize to other domains. 
%We have shown a general method for generating hard initial states for search problems. 
This is because all techniques used in our system are Sokoban independent. %Backward search is a general algorithm that is used in many problems.
%Novelty is a domain-independent technique with wide spread success in the task of solving problems in a diverse set of formalisms.
%PDB heuristics were originally proposed as domain-dependent techniques. However, are now one of the most effective domain-independent heuristic functions. 
Moreover, Sokoban is a prototypical problem of a large class of problems known as transportation problems. % which is the most common class of planning problems. 
Problems in this class has four central characteristics: there is a set of connected locations, a set of movable objects and a set of goal locations, and the solution is a sequence of actions that move (a subset of) the movable objects to (a subset of) the goal locations. Sokoban presents all these characteristics thus our system is likely to perform well in these problems as well.

\section{Conclusions}

In this paper we presented a system for generating hard and solvable initial states of Sokoban. Our system~$\beta$ performs a backward search from the goal states of a problem and selects states using PDB-based hardness metrics. 
We also proposed the use of novelty to guide $\beta$'s search. 
%We showed empirically that our system 
%outperforms existing methods for generating initial states of Sokoban. It 
%generates initial states with an order of magnitude more boxes than states generated by previous methods. 
%This large improvement on the size of problems our system generates will potentially allow learning algorithms that depend on generative systems to scale to larger problems. 
Compared to the states designed by human experts, $\beta$ is able to generate states that are harder to solve by a solver. $\beta$ is the first generative system of initial states of Sokoban with superhuman performance in terms of difficulty for a solver. 
%As future works, we plan to apply $\beta$ to other problem domains.

\section*{Acknowledgments}
André Pereira acknowledges support from FAPERGS with project $17/2551-0000867-7$ and Levi Lelis acknowledges support from FAPEMIG and CNPq.
This study was financed in part by the Coordenação de Aperfeiçoamento de Pessoal de Nível Superior - Brasil (CAPES) - Finance Code 001.

%\appendix
\clearpage
\newpage
\footnotesize
\bibliographystyle{named}
\bibliography{ijcai19}

\end{document}